\documentclass[10pt,twocolumn,letterpaper]{article}
\usepackage[accsupp]{axessibility}
\usepackage[pagenumbers]{cvpr} 
\usepackage{booktabs}
\usepackage{multirow}
\usepackage{makecell}
\usepackage{float}

\newcommand{\subtitle}[1]{{\noindent}{\textbf{#1}}}

\usepackage[ruled]{algorithm2e} 
\usepackage{algorithmic}
\definecolor{cvprblue}{rgb}{0.21,0.49,0.74}
\usepackage[pagebackref,breaklinks,colorlinks,allcolors=cvprblue]{hyperref}

\def\paperID{} 
\def\confName{CVPR}
\def\confYear{2026}

\title{ZeroIDIR: Zero-Reference Illumination Degradation Image Restoration with Perturbed Consistency Diffusion Models}

\author{
	Hai Jiang\textsuperscript{\rm1}, Zhen Liu\textsuperscript{\rm2}, Yinjie Lei\textsuperscript{\rm3},  Songchen Han\textsuperscript{\rm1}, Bing Zeng\textsuperscript{\rm2}, Shuaicheng Liu\textsuperscript{\rm2,$^\dagger$} \\
	\textsuperscript{\rm1}School of Aeronautics and Astronautics, Sichuan University \\ 
	\textsuperscript{\rm2}University of Electronic Science and Technology of China \\
	\textsuperscript{\rm3}College of Electronics and Information Engineering, Sichuan University \\
	{\tt\small jianghai@stu.scu.edu.cn, \{liuzhen03@std., liushuaicheng@\}uestc.edu.cn}
}

\begin{document}
\twocolumn[{%
    \maketitle
    \begin{figure}[H]
    \hsize=\textwidth
    \centering
    \includegraphics[width=2.1\linewidth]{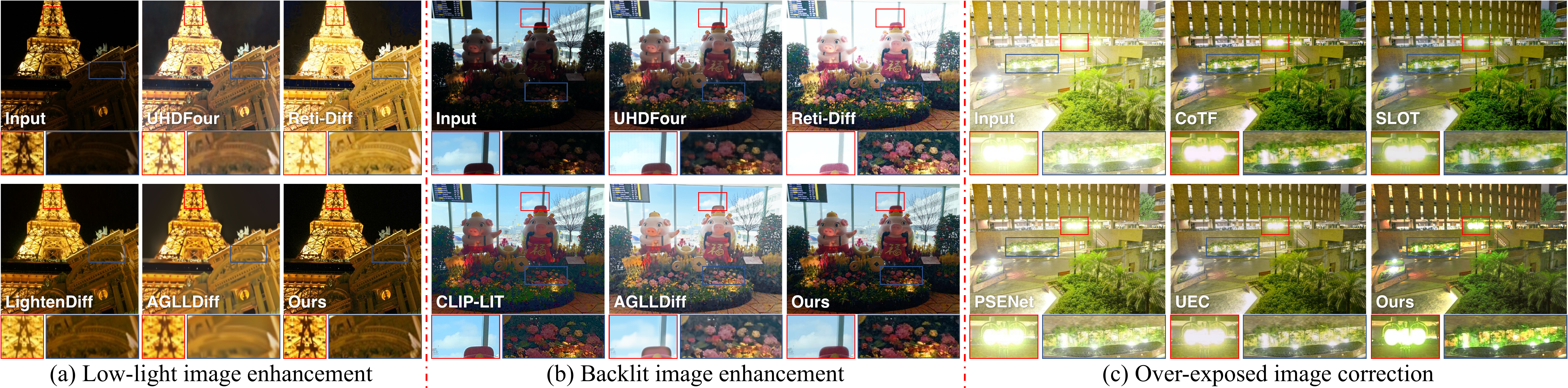} 
    \caption{Visual comparisons of our method with recent state-of-the-art supervised IDIR methods UHDFour~\cite{UHD_ICLR}, AnlightenDiff~\cite{Anlightendiff}, CoTF~\cite{CoTF}, Reti-Diff~\cite{Retidiff}, and SLOT~\cite{Exposure-slot}, as well as unsupervised methods PSENet~\cite{Psenet}, CLIP-LIT~\cite{CLIPLIT}, LightenDiff~\cite{Lightendiffusion}, UEC~\cite{UEC}, and AGLLDiff~\cite{AGLLDiff}. Previous methods appear to have incorrect exposure, blurred details, or noise amplification to degrade visual quality, while our method properly corrects global and local illumination and reconstructs sharper details.}
    \label{fig: teaser}
\end{figure}
}]
\renewcommand{\thefootnote}{\fnsymbol{footnote}}
\footnotetext[2]{Corresponding author.}

\maketitle
\begin{abstract}
In this paper, we propose a zero-reference diffusion-based framework, named ZeroIDIR, for illumination degradation image restoration, which decouples the restoration process into adaptive illumination correction and diffusion-based reconstruction while being trained solely on low-quality degraded images. Specifically, we design an adaptive gamma correction module that performs spatially varying exposure correction to generate illumination-corrected only representations to mitigate exposure bias and serve as reliable inputs for subsequent diffusion processes, where a histogram-guided illumination correction loss is introduced to regularize the corrected illumination distribution toward that of natural scenes. Subsequently, the illumination-corrected image is treated as an intermediate noisy state for the proposed perturbed consistency diffusion model to reconstruct details and suppress noise. Moreover, a perturbed diffusion consistency loss is proposed to constrain the forward diffusion trajectory of the final restored image to remain consistent with the perturbed state, thus improving restoration fidelity and stability in the absence of supervision. Extensive experiments on publicly available benchmarks show that the proposed method outperforms state-of-the-art unsupervised competitors and is comparable to supervised methods while being more generalizable to various scenes. Code is available at \url{https://github.com/JianghaiSCU/ZeroIDIR}.
\end{abstract}
   
\section{Introduction}\label{sec: intro}
Images captured under challenging illumination conditions, such as low-light, backlit, and under-/over-exposed scenes, often suffer from various degradations including poor visibility and noise, which significantly hinder the performance of downstream vision tasks~\cite{Classification, low_light_detection}. To restore such degraded low-quality images, extensive research efforts have been devoted over the past decades. Early approaches~\cite{SRIE, LIME} mainly relied on hand-crafted priors, such as histogram equalization~\cite{HE} and Retinex theory~\cite{Retinex} for illumination correction and detail recovery. However, selecting suitable priors for diverse illumination conditions remains challenging, as illumination degradation image restoration (IDIR) is inherently ill-posed, thus restricting the practical application of these methods.

These limitations have been partially alleviated by the rise of deep learning. Learning-based methods~\cite{SNRNet, SMG, GCCIM, Retinexformer, RAVE, Hqg-net} leverage well-designed architectures and training strategies to directly map degraded inputs to normal-light images, demonstrating stronger robustness than traditional techniques. Despite remarkable progress, these methods often suffer from overfitting and limited generalization to real-world scenes, leading to visually unsatisfactory results. As shown in Fig.~\ref{fig: teaser}, previous state-of-the-art learning-based supervised methods UHDFour~\cite{UHD_ICLR}, CoTF~\cite{CoTF}, and SLOT~\cite{Exposure-slot}, as well as unsupervised methods CLIP-LIT~\cite{CLIPLIT}, PSENet~\cite{Psenet}, and UEC~\cite{UEC} present incorrect exposure, blurred details, and amplified noise.

Recently, generative approaches, especially diffusion models~\cite{ddpm, ddim}, have been introduced into IDIR to enhance perceptual quality. Most existing diffusion-based methods~\cite{PyDiff, DiffLL, Diff-Retinex++, GSAD, Retidiff} rely on large-scale paired datasets under the conditional framework~\cite{conditional_ddpm} for supervised training, enabling favorable illumination correction and detail reconstruction while the learned distribution is inevitably constrained by the paired data. As shown in Fig.~\ref{fig: teaser}, the supervised diffusion-based method Reti-Diff~\cite{Retidiff} produces over-exposed results with noise amplification. To overcome the reliance on paired data, unsupervised diffusion models have recently been explored. On the one hand, zero-shot solutions~\cite{AGLLDiff, FourierDiff, GDP} leverage the rich priors of pre-trained diffusion models, thus avoiding training from scratch. On the other hand, several studies~\cite{Lightendiffusion, difflle} investigate training with unpaired data, where degraded images are aligned with unpaired normal-light references to guide restoration. While these unsupervised approaches alleviate the reliance on paired datasets, they are still limited by either the restricted degradation types assumed in zero-shot settings or the distribution mismatch in unpaired data, resulting in inferior performance under real-world scenes. As shown in Fig.~\ref{fig: teaser}, the zero-shot method AGLLDiff~\cite{AGLLDiff} and the unpaired training method LightenDiff~\cite{Lightendiffusion} produce smoothed details, color distortion, and unstable exposure correction. Moreover, diffusion models, despite their superior high-frequency detail information generation capability, tend to be biased in low-frequency generation, particularly in exposure~\cite{diffusion_bias}, posing challenges for jointly addressing illumination correction and detail restoration in IDIR.

To this end, we propose ZeroIDIR, a zero-reference diffusion-based framework for illumination degradation image restoration, encompassing low-light image enhancement (LLIE), backlit image enhancement (BIE), and multiple exposure correction (MEC). ZeroIDIR decouples the restoration process into adaptive illumination correction and diffusion-based reconstruction, enabling visually faithful results trained solely on degraded images without paired supervision or external references. Specifically, we propose an adaptive gamma correction module (AGCM) that performs spatially varying exposure correction while preserving structural content and avoiding noise amplification to prepare illumination-corrected only results for subsequent restoration. A histogram-guided illumination correction loss further regularizes the corrected illumination distribution toward that of natural scenes. Subsequently, the refined image serves as an intermediate noisy state for our perturbed consistency diffusion model (PCDM), which leverages its generative prior and inherent denoising capability to generate the final high-quality image, where a perturbed diffusion consistency loss is designed to align the diffusion trajectory of the restored image with the input illumination-corrected image, thereby enabling reliable fine-grained detail reconstruction and noise suppression. As shown in Fig.~\ref{fig: teaser}, ZeroIDIR effectively restores global and local illumination and reconstructs sharper details. 

Our contributions can be summarized as follows:
\begin{itemize}
    \item We propose a zero-reference diffusion-based framework, termed ZeroIDIR, for illumination degradation image restoration. Extensive experiments demonstrate that our method outperforms existing state-of-the-art unsupervised competitors while being comparable and having better generalization abilities than supervised methods.
    \item We introduce an adaptive gamma correction module to perform spatially varying exposure correction to prepare illumination-corrected results, where a histogram-guided illumination correction loss is introduced to regularize the illumination distribution toward that of natural scenes.
    \item We design a perturbed consistency diffusion model that takes the illumination-corrected image as the intermediate noisy state for detail reconstruction and noise suppression, where a perturbed diffusion consistency loss is proposed to improve restoration fidelity and stability.
\end{itemize}
\begin{figure*}[!t]
    \centering
    \includegraphics[width=\linewidth]{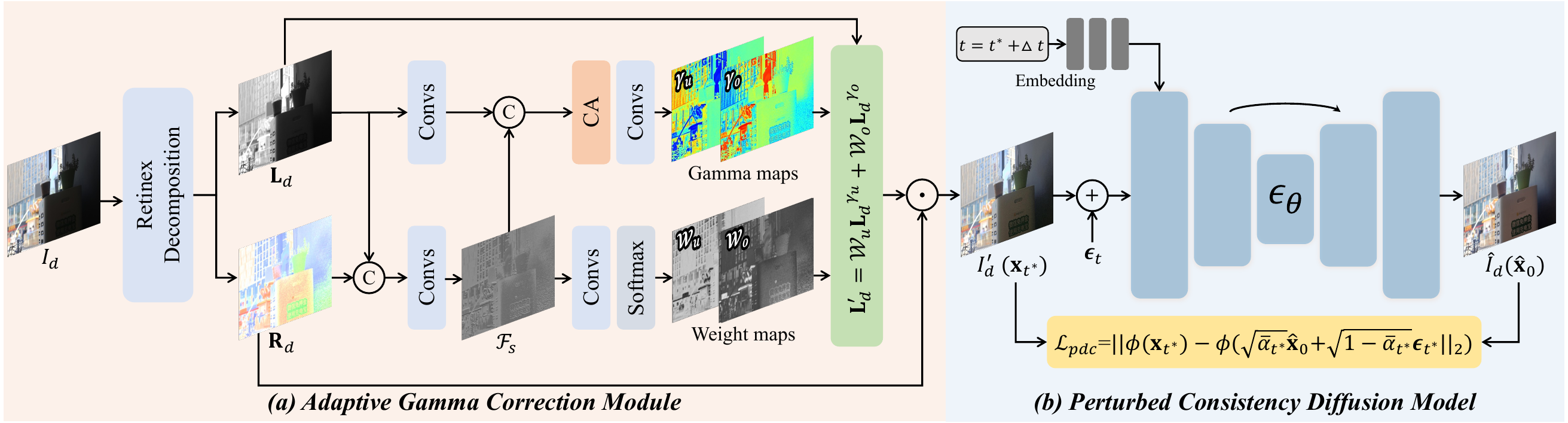}
    \caption{The overall pipeline of our proposed ZeroIDIR framework. We first perform Retinex decomposition on the illumination degradation image $I_{d}$ to generate a reflectance map $\mathbf{R}_{d}$ and an illumination map $\mathbf{L}_{d}$. Then, we design an adaptive gamma correction module (AGCM) to achieve spatially adaptive illumination correction on the $\mathbf{L}_{d}$ through two dedicated branches to prepare an illumination-corrected only image $I_{d}^{'}$ for subsequent reconstruction. The refined image is then treated as the intermediate perturbed noisy state $\mathbf{x}_{t^{\ast}}$ and fed into a perturbed consistency diffusion model (PCDM) for reconstruction, with the guidance of the proposed perturbed diffusion consistency loss $\mathcal{L}_{pdc}$ toward the final restored sharp image $\hat{I}_{d}$ in a zero-reference manner.}
    \label{fig: Pipeline}
\end{figure*}

\section{Related Work}
\subsection{Illumination Degradation Image Restoration}\label{subsec: related_work_IDIR}

\subtitle{Low-light image enhancement (LLIE)} methods were initially dominated by traditional handcrafted techniques~\cite{HE, Retinex}. With the advent of deep learning, supervised methods~\cite{MIRNet, SNRNet, GCCIM, RFLLIE} have achieved remarkable progress while highly relying on large-scale paired datasets and thus suffer from limited generalization ability. Unsupervised methods~\cite {Zero-DCE, EnlightenGAN, RUAS, SCI, PairLIE, NeRCo} utilize their characteristics of not requiring paired data to solve the LLIE by employing adversarial learning, curve estimation, or neural architecture search with better generalization ability. 

\subtitle{Backlit image enhancement (BIE)} shares similarities with LLIE but emphasizes reconstructing under-exposed foregrounds against well-exposed backgrounds. In analogy to LLIE, traditional methods~\cite{BIE_traditional1, BIE_traditional2} often presents artifacts or over-amplification due to their reliance on hand-crafted priors. To facilitate learning-based methods, the first paired dataset BAID~\cite{BAID} was introduced, where ground truth images are manually adjusted by professional photographers to ensure high-quality references. Moreover, inspired by pre-trained vision-language models, several researchers~\cite{CLIPLIT, RAVE} explore the potential of CLIP~\cite{CLIP} to produce visually appealing and context-aware results.

\subtitle{Multiple exposure correction (MEC)} tends to address both under-exposure and over-exposure simultaneously. Earlier works~\cite{MEC1, MEC2} tackled single-type exposure errors, but unified frameworks have since been explored to handle mixed degradations. Supervised methods~\cite{MSEC, CMEC, CoTF, Exposure-slot, MSLT, FECNet} either create dense translations between input and output pairs or predict mapping curve parameters to restore images. However, these works for general exposure correction are excessively dependent on training data, leading to poor generalization. In contrast, unsupervised methods~\cite{PEC, Exposure, Dual, UEC, ExCNet, Psenet} employ exposure-sensitive compensation, radiometry modeling, or pseudo label construction to eliminate biases from the paired data and consequently improve generalization ability.

\subsection{Diffusion-base Image Restoration}\label{subsec: diffusion_IR}
With the rapid progress of diffusion models in low-level vision~\cite{Survey, RAWFlow, Solving, Ispdiffuser, Foundir, DMHomo, rolling_shutter,StableMotion,denoising_diffusion,rolling_shutter_2,video_deblur,RecDiffusion,RealSH_plus, UnfoldLDM}, a variety of image restoration tasks have been explored, especially illumination degradation image restoration~\cite{DiffLL, Retidiff, Unfoldir, sied, DiffFourier}. Most existing methods rely on the conditional mechanism~\cite{conditional_ddpm}, where models are trained from scratch with paired data. To alleviate paired data dependence, several methods~\cite{difflle, Lightendiffusion} adopt unpaired data to optimize diffusion models, while being inherently vulnerable to the domain gap between the unpaired data. Moreover, some methods~\cite{GDP, FourierDiff, AGLLDiff} employ zero-shot strategies that leverage the priors from the pre-trained diffusion models to restore degraded images without reference images. However, their performance is hampered by the pre-trained models, leading to restored results with unsatisfactory visual quality. In this paper, we propose a zero-reference framework that performs illumination correction and diffusion-based reconstruction separately to achieve visually satisfactory IDIR.

\section{Method}\label{sec: method}
\subsection{Overview}\label{subsec: overview}
The overall pipeline of our proposed ZeroIDIR is illustrated in Fig.~\ref{fig: Pipeline}. Given an illumination-degraded image $I_{d}$, we first perform Retinex decomposition to separate it into a reflectance map and an illumination map. Based on these components, we design an adaptive gamma correction module (AGCM) that estimates spatially varying gamma maps and corresponding weight maps through two dedicated branches, aiming to generate an illumination-corrected only representation $I^{'}_{d}$. Subsequently, the refined image is fed into our perturbed consistency diffusion model (PCDM) as a pseudo intermediate noisy state, where the proposed perturbed diffusion consistency loss is adopted to guide the reconstruction processes toward the final visually satisfactory image $\hat{I}_{d}$ in a zero-reference manner.

\subsection{Adaptive Gamma Correction Module}\label{subsec: agcm}
Illumination degradation image restoration involves three critical concerns: exposure correction, detail reconstruction, and noise suppression. Recently, diffusion models have attracted considerable attention due to their impressive generative capability. However, they often suffer from low-frequency generative bias, particularly in exposure~\cite{diffusion_bias}. To this end, we propose an adaptive gamma correction module (AGCM) that performs spatially varying exposure correction before the diffusion stage, generating illumination-corrected representations to serve as reliable inputs for subsequent reconstruction while mitigating exposure bias.

Specifically, as shown in Fig.~\ref{fig: Pipeline}(a), we first decompose the illumination-degraded image $I_{d}$ into a reflectance map $\textbf{R}_{d}$ and an illumination map $\textbf{L}_{d}$, where $\textbf{R}_{d}$ represents the inherent content information that should be consistent under diverse illumination conditions and $\textbf{L}_{d}$ indicates the contrast and brightness information. We then operate on the illumination component rather than the original image to achieve spatially adaptive illumination correction while maintaining the inherent content information unchanged and avoiding noise amplification. The $\textbf{L}_{d}$ is concatenated with $\textbf{R}_{d}$ and processed through convolutional blocks to construct an illumination-aware feature $\mathcal{F}_{s}$ that provides structural guidance. In parallel, $\mathbf{L}_{d}$ itself serves as a global exposure prior and is fused with the structure feature through channel-attention (CA) module and several convolutional blocks to predict two spatially varying gamma maps $\{\gamma_{u}, \gamma_{o}\} \in \mathbb{R}^{H \times W \times 1}$ that perform exposure correction for under-exposed and over-exposed regions, respectively. To adaptively balance these two corrections, a pair of spatial weight maps $\{\mathcal{W}_{u}, \mathcal{W}_{o}\} \in \mathbb{R}^{H \times W \times 1}$ are predicted from $\mathcal{F}_{s}$ to determine the relative contributions of the two gamma-corrected results. The corrected illumination map $\mathbf{L}^{'}_{d}$ is therefore obtained as:
\begin{equation}
    \mathbf{L}^{'}_{d} = \mathcal{W}_{u} {\textbf{L}_{d}}^{\gamma_{u}} + \mathcal{W}_{o} {\textbf{L}_{d}}^{\gamma_{o}},
\end{equation}
and subsequently combined with the reflectance map to yield the illumination-corrected result $I^{'}_{d} = \mathbf{L}^{'}_{d} \odot \mathbf{R}_{d}$, where $\odot$ denotes the Hadamard product operation. 

\begin{figure}[!t]
    \centering
    \includegraphics[width=\linewidth]{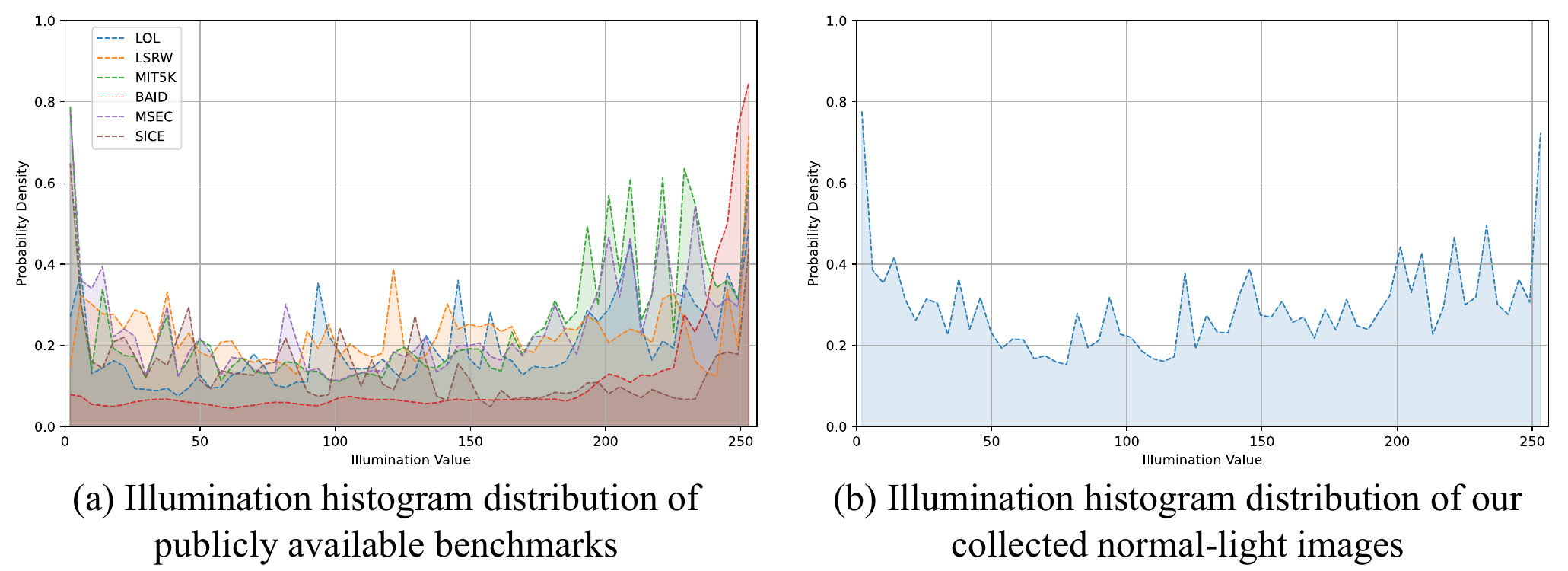}
    \caption{The illumination histogram distributions derived from the normal-light images in publicly available benchmarks~\cite{RetinexNet, R2RNet, MIT5K, BAID, MSEC, SICE} (a) and our collected $\sim20k$ normal-light images (b).}
    \label{fig: hist}
\end{figure}

\subtitle{Histogram-Guided Illumination Correction Loss.} To enable adaptive illumination correction within AGCM, we introduce a histogram-guided illumination correction loss $\mathcal{L}_{hic}$ that constrains the illumination distribution of the corrected image to align with that observed in natural scenes. As illustrated in Fig.~\ref{fig: hist}, we first analyze the histogram distributions of illumination maps derived from reference normal-light images in several widely used benchmarks~\cite{RetinexNet, R2RNet, MIT5K, SICE, MSEC}, where these histograms exhibit a stable and concentrated distribution pattern, reflecting the natural exposure characteristics of well-illuminated environments. Motivated by this observation, we further collect a large corpus of normal-light images and aggregate their histograms to form an empirical prior distribution. Accordingly, $\mathcal{L}_{hic}$ encourages the exposure statistics of illumination-corrected results to follow this prior as:
\begin{equation}\label{eq: hic}
    \mathcal{L}_{hic} = D_{KL}(\mathcal{H}(\mathbf{L}^{'}_{d})\Big\|\mathbf{H}_{\text{prior}}),
\end{equation}
where $\mathcal{H}(\cdot)$ denotes the histogram computation operator, and $\mathbf{H}_{\text{prior}}$ represents the empirical prior distribution estimated from real-world well-illuminated images.

\subsection{Perturbed Consistency Diffusion Model}\label{subsec: pcdm}
While AGCM produces illumination-enhanced results with natural brightness, it still suffers from detail loss and noise interference. To address this, we propose a perturbed consistency diffusion model (PCDM) that leverages the generative ability and intrinsic denoising capability of diffusion models to reconstruct high-quality images. 

In standard diffusion frameworks~\cite{ddim, ddpm}, a clean image $\mathbf{x}_{0}$ is progressively perturbed into Gaussian noise over $T$ steps through the forward diffusion process as:
\begin{equation}
    \mathbf{x}_t=\sqrt{\bar{\alpha}_t} \mathbf{x}_0+\sqrt{1-\bar{\alpha}_t} \boldsymbol{\epsilon}_t,
\end{equation}
where $\mathbf{x}_{t}$ indicates the corrupted noisy data at time-step $t\in[0,T]$ and $\boldsymbol{\epsilon}_t \sim \mathcal{N}(\mathbf{0},\mathbf{I})$. In our zero-reference setting, however, the clean target $\mathbf{x}_{0}$ is not accessible. We therefore reinterpret the illumination-corrected results $I^{'}_{d}$ from AGCM as a partially diffused version of its unknown, high-quality counterpart, corresponding to the diffusion state at time step $t^{\ast}$, denoted as $\mathbf{x}_{t^{\ast}}$. This formulation enables direct embedding of the intermediate illumination-corrected noisy image into the diffusion trajectory without requiring paired supervision. To construct valid training samples, we further continue the forward diffusion process from $\mathbf{x}_{t^{\ast}}$ by injecting additional noise over $\Delta t$ steps as:
\begin{equation}
    \mathbf{x}_t=\sqrt{\frac{\bar{\alpha}_t}{\bar{\alpha}_t^{\ast}}}\mathbf{x}_{t^{\ast}} + \sqrt{1-\frac{\bar{\alpha}_t}{\bar{\alpha}_t^{\ast}}}\boldsymbol{\epsilon}_t, t=t^{\ast}+\Delta t,
\end{equation}
where $t^{\ast}$ and $\Delta t$ are randomly sampled within predefined bounds, aiming to facilitate the diffusion model learning to handle various degradation intensities.

In the training phase, the objective of the diffusion model is to optimize the parameters $\theta$ of the denoising network $\boldsymbol{\epsilon}_\theta$ so that the estimated noise $\boldsymbol{\epsilon}_\theta(\mathbf{x}_{t}, t, \mathbf{y})$ approximates the sampled Gaussian noise like~\cite{ddpm}, which is formulated as:
\begin{equation}
    \mathcal{L}_{diff} = ||\boldsymbol{\epsilon}_t - \boldsymbol{\epsilon}_\theta(\mathbf{x}_{t}, t, \mathbf{y})||_{2}.
\end{equation}
Here, $\mathbf{y}$ denotes the intermediate illumination-corrected result instead of the original low-quality image, which serves as the condition and allows the model to focus on fine-grained detail reconstruction and noise suppression. 
\begin{algorithm}[!t]
\caption{PCDM training}
\label{alg: pcdm}
\KwIn{The illumination-corrected image $I^{'}_{d}$, time step $T$, noise prediction network $\boldsymbol{\epsilon}_\theta$.

$\mathbf{x}_{t^\ast}$ = $I^{'}_{d}$, $\mathbf{y}$ = $I^{'}_{d}$}

\KwOut{Trained parameters $\theta$}

\While{not converged}{
    $t = t^\ast + \Delta t$, $\Delta t \sim \mathcal{U}(t^\ast, T - t^\ast)$\\[2pt]

    Sampling $\boldsymbol{\epsilon}_t \sim \mathcal{N}(\mathbf{0}, \mathbf{I})$, $\boldsymbol{\epsilon}_{t^\ast} \sim \mathcal{N}(\mathbf{0}, \mathbf{I})$\\[2pt]
    
    $\mathbf{x}_t = \sqrt{\frac{\bar{\alpha}_t}{\bar{\alpha}_{t^\ast}}} \mathbf{x}_{t^\ast} + \sqrt{1 - \frac{\bar{\alpha}_t}{\bar{\alpha}_{t^\ast}}} \boldsymbol{\epsilon}_t$ \\[2pt]
   
    \text{Perform gradient descent steps on}
    $\nabla_\theta\left\|\boldsymbol{\epsilon}_t - \boldsymbol{\epsilon}_\theta(\mathbf{x}_t, t, \mathbf{y})\right\|_2$\\[2pt]
    
    $\hat{\mathbf{x}}_0 = \frac{1}{\sqrt{\bar{\alpha}_t}} \left( \mathbf{x}_t - \sqrt{1 - \bar{\alpha}_t} \boldsymbol{\epsilon}_\theta(\mathbf{x}_t, t, \mathbf{y}) \right)$\\[2pt]
    
    $\hat{\mathbf{x}}_{t^\ast} = \sqrt{\bar{\alpha}_{t^\ast}} \hat{\mathbf{x}}_0 + \sqrt{1 - \bar{\alpha}_{t^\ast}} \boldsymbol{\epsilon}_{t^\ast}$\\[2pt]

    \text{Perform gradient descent steps on}
    $\nabla_\theta\left\|\phi(\mathbf{x}_{t^\ast}) - \phi(\hat{\mathbf{x}}_{t^\ast}) \right\|_2$
}
\end{algorithm}

\subtitle{Perturbed Diffusion Consistency Loss.} Moreover, to encourage the model to generate final high-quality images $\hat{I}_{d}$, i.e., $\hat{\mathbf{x}}_{0}$, whose forward diffusion trajectory remains faithful to the intermediate state, we propose a perturbed diffusion consistency loss $\mathcal{L}_{pdc}$ to enforce consistency in the feature space between the illumination-corrected input $\mathbf{x}_{t^\ast}$ and the reconstructed state $\hat{\mathbf{x}}_{t^\ast}$ that is produced by passing the generated clean $\hat{\mathbf{x}}_{0}$ through the forward diffusion process for $t^{\ast}$ steps. 
The $\mathcal{L}_{pdc}$ is therefore formulated as:
\begin{equation}\label{eq: pdc}
    \mathcal{L}_{pdc} = ||\phi(\mathbf{x}_{t^\ast}) - \phi(\sqrt{\bar{\alpha}_t^{\ast}}\hat{\mathbf{x}}_{0}+\sqrt{1-\bar{\alpha}_t^{\ast}}\boldsymbol{\epsilon}_{t^{\ast}})||_2,
\end{equation}
where $\hat{\mathbf{x}}_{0}$ is estimated from the predicted noise vector following~\cite{QuadPrior} and $\phi(\cdot)$ denotes the pre-trained VGG-16
model. Overall, the training strategy of the proposed PCDM is summarized in Algorithm~\ref{alg: pcdm}.
\begin{table*}[!t]
  \centering
  \caption{Quantitative comparisons on the LOL~\cite{RetinexNet}, LSRW~\cite{R2RNet}, and MIT5K~\cite{MIT5K} datasets for LLIE. The best results are highlighted in \textbf{bold} and the second-best are in \underline{underlined}. `SL' and `UL' denote the methods as supervised and unsupervised, respectively.}
  \resizebox{\linewidth}{!}{
    \begin{tabular}{c|l|c|ccc|ccc|ccc}
    \toprule
    \multirow{2}[4]{*}{Type} & \multirow{2}[4]{*}{Methods} & \multicolumn{1}{c|}{\multirow{2}[4]{*}{Reference}} & \multicolumn{3}{c|}{LOL~\cite{RetinexNet}} & \multicolumn{3}{c|}{LSRW~\cite{R2RNet}} & \multicolumn{3}{c}{MIT5K~\cite{MIT5K}} \\
    \cmidrule{4-12} & & & PSNR $\uparrow$ & SSIM $\uparrow$ & LPIPS $\downarrow$ & PSNR $\uparrow$ & SSIM $\uparrow$ & LPIPS $\downarrow$ & PSNR $\uparrow$ & SSIM $\uparrow$ & LPIPS $\downarrow$ \\
    \midrule
    \multirow{6}[2]{*}{SL} 
    & RetinexNet~\cite{RetinexNet} & BMVC' 18 & 16.774 & 0.462 & 0.390 & 15.609 & 0.414 & 0.393 & 11.268 & 0.647 & 0.303 \\
    & KinD++~\cite{KinD++} & IJCV' 21 & 17.752 & 0.758 & 0.198 & 16.085 & 0.394 & 0.366 & 14.158 & 0.679 & 0.209 \\
    & UHDFour~\cite{UHD_ICLR} & ILCR' 23 & 23.093 & 0.821 & 0.142 & 17.300 & 0.529 & 0.362 & 17.207 & 0.543 & 0.234 \\
    & PyDiff~\cite{PyDiff} & IJCAI' 23 & \underline{23.275} & \underline{0.859} & \underline{0.108} & 17.264 & 0.510 & 0.335 & 18.617 & 0.792 & 0.189 \\
    & AnlightenDiff~\cite{Anlightendiff} & TIP' 24 & 19.834 & 0.748 & 0.178 & 18.016 & 0.501 & 0.351 & 16.676 & 0.716 & 0.238 \\
    & Reti-Diff~\cite{Retidiff} & ICLR' 25 & \textbf{25.180} & \textbf{0.869} & \textbf{0.094} & 16.441 & 0.505 & \underline{0.304} & 13.372 & 0.740 & 0.187 \\
    \midrule
    \multirow{11}[2]{*}{UL} 
    & Zero-DCE~\cite{Zero-DCE} & CVPR' 20 & 14.861 & 0.562 & 0.330 & 15.867 & 0.443 & 0.315 & 13.500 & 0.699 & 0.235 \\
    & EnlightenGAN~\cite{EnlightenGAN}    & TIP' 21 & 17.606 & 0.653 & 0.319 & 17.106 & 0.463 & 0.322 & 13.275 & 0.728 & 0.205 \\
    & RUAS~\cite{RUAS}  & CVPR' 21 & 16.405 & 0.503 & 0.257 & 14.271 & 0.461 & 0.455 & 5.168 & 0.360 & 0.682 \\
    & SCI~\cite{SCI}   & CVPR' 22 & 14.784 & 0.525 & 0.333 & 15.242 & 0.419 & 0.321 & 7.828 & 0.568 & 0.368 \\
    & PairLIE~\cite{PairLIE} & CVPR' 23 & 19.514 & 0.731 & 0.254 & 17.602 & 0.501 & 0.323 & 9.253 & 0.608 & 0.316 \\
    & NeRCo~\cite{NeRCo} & ICCV' 23 & 19.738 & 0.740 & 0.239 & 17.844 & 0.535 & 0.371 & 17.319 & 0.770 & 0.217 \\
    & FourierDiff~\cite{FourierDiff} & CVPR' 24 & 17.557 & 0.612 & 0.277 & 15.643 & 0.458 & 0.322 & 17.812 & 0.794 & \underline{0.160} \\
    & QuadPrior~\cite{QuadPrior} & CVPR' 24 & 18.771 & 0.777 & 0.206 & 16.958 & \underline{0.552} & 0.403 & 18.147 & 0.761 & 0.194 \\
    & LightenDiff~\cite{Lightendiffusion} & ECCV' 24 & 20.190 & 0.809 & 0.182 & \underline{18.388} & 0.525 & 0.313 & \textbf{21.248} & \underline{0.799} & 0.181 \\
    & AGLLDiff~\cite{AGLLDiff} & AAAI' 25 & 19.836 & 0.806 & 0.192 & 17.359 & 0.544 & 0.359 & 8.938 & 0.622 & 0.316 \\
    & ZeroIDIR (Ours)  & -     & 20.874 & 0.811 & 0.167 & \textbf{18.823} & \textbf{0.563} & \textbf{0.301} & \underline{20.327} & \textbf{0.806} & \textbf{0.151} \\
    \bottomrule
    \end{tabular}}
  \label{tab: LLIE_compare}%
\end{table*}

\subsection{Network Training}\label{subsec: network_training}
Our approach adopts a two-stage strategy for network training. We collect $10k$ low-quality illumination degradation images, including low-light images, backlit images, and over-exposed images, to comprise the training data. In the first stage, we aim to optimize the AGCM while freezing the parameters of the diffusion model for illumination correction through Eq.(\ref{eq: hic}). Moreover, we follow~\cite{Zero-DCE} to employ the exposure control loss $\mathcal{L}_{exp}$ to guide the exposure intensity of the illumination-corrected images toward that of  the well-exposedness level $E$, which is formulated as:
\begin{equation}
    \mathcal{L}_{exp} = \frac{1}{M}\sum_{k=1}^{M}||E_{k} - E||_1,
\end{equation}
where $M$ denotes the number of nonoverlapping local regions of size $16\times16$ and $E$ is set as 0.6 following~\cite{Zero-DCE}. $E_{k}$ denotes the average intensity value of a local region in the illumination-corrected image. We also employ the edge-aware total variation loss $\mathcal{L}_{eatv}$ to regularize the predicted gamma maps to ensure stable and structure-preserving illumination correction, which is formulated as:
\begin{equation}
    \mathcal{L}_{eatv} = \sum_{i \in \{u,o\}}||\nabla\gamma_{i}\cdot \text{exp}(-\lambda_g\nabla\mathbf{R}_{d})||_2,
\end{equation}
where $\nabla$ denotes the horizontal and vertical gradients, and $\lambda_g$ is the coefficient to balance the perceived strength of the structure. The overall training objective function used to optimize the AGCM is formulated as $\mathcal{L}_{stage1} = \mathcal{L}_{exp} + \lambda_1 \mathcal{L}_{hic} + \lambda_2 \mathcal{L}_{eatv}$. In the second stage, we optimize the PCDM through $\mathcal{L}_{stage2}=\mathcal{L}_{diff} + \lambda_3 \mathcal{L}_{pdc}$ while freezing the model parameters of the AGCM.

\section{Experiments}\label{sec: experiments}
\subsection{Experimental Settings}\label{subsec:experimental_settings}
\subtitle{Implementation Details.} We implement the proposed method with PyTorch on one NVIDIA A100 GPU, where the batch size and patch size are set to 4 and 256 $\times$256. The networks can be converged after training in two stages with $1 \times 10^5$ and $1 \times 10^6$ iterations, respectively. We employ the Adam optimizer~\cite{Adam} for optimization with the initial learning rate set to $1 \times 10^{-4}$ in the first stage and decays by a factor of 0.8 while reinitializing it to a fixed value of $8 \times 10^{-5}$ in the second stage. The hyperparameters $\lambda_{1}$, $\lambda_{2}$, $\lambda_{3}$, and $\lambda_{g}$ are empirically set to 0.5, 0.1, 1.0, and 20.0, respectively. For our PCDM, the U-Net~\cite{Unet} architecture is adopted as the noise estimator network with the time step $T$ set to 1000 for the forward diffusion and the sampling step set to 20 for the reverse denoising process. The $t^{\ast}$ is predefined within the range $[0, 50]$ and $\Delta t \sim \mathcal{U}(t^\ast, T - t^{\ast})$.

\subtitle{Datasets and Metrics.} For the low-light image enhancement task, we conduct experiments on three paired datasets, including LOL~\cite{RetinexNet}, LSRW~\cite{R2RNet}, and MIT5K~\cite{MIT5K}. For the backlit image enhancement task, we conduct experiments on the paired BAID~\cite{BAID} dataset and the real-world unpaired benchmark Backlit300~\cite{CLIPLIT}. In the multiple exposure correction task, we conduct experiments on two paired datasets that contain paired under-/over-exposed images and well-exposed images, including MSEC~\cite{MSEC} and SICE~\cite{SICE}. For paired datasets, we adopt two distortion metrics PSNR and SSIM~\cite{SSIM}, and a full-reference perceptual metric LPIPS~\cite{LPIPS} for evaluation. For the unpaired dataset, we use two non-reference perceptual metrics NIQE~\cite{NIQE} and CLIPIQA~\cite{CLIPIQA} to measure the visual quality.

\subsection{Quantitative Comparison}\label{subsec: Quantitative_Comparison}
\subtitle{Low-light Image Enhancement.} For the low-light image enhancement task, we compare the proposed method with existing state-of-the-art supervised methods including RetinexNet~\cite{RetinexNet}, KinD++~\cite{KinD++}, UHDFour~\cite{UHD_ICLR}, PyDiff~\cite{PyDiff}, AnlightenDiff~\cite{Anlightendiff}, and Reti-Diff~\cite{Retidiff}, as well as unsupervised methods including Zero-DCE~\cite{Zero-DCE}, EnlightenGAN~\cite{EnlightenGAN}, RUAS~\cite{RUAS}, SCI~\cite{SCI}, PairLIE~\cite{PairLIE}, NeRCo~\cite{NeRCo}, FourierDiff~\cite{FourierDiff}, QuadPrior~\cite{QuadPrior}, LightenDiff~\cite{Lightendiffusion}, and AGLLDiff~\cite{AGLLDiff}. For fair comparisons, we adopt the released weights pre-trained on the LOL training set of supervised methods for evaluation. As reported in Table~\ref{tab: LLIE_compare}, our method outperforms all unsupervised competitors except for the suboptimal PSNR on the MIT5K dataset. The reason we cannot surpass supervised approaches on the LOL dataset is that they are typically trained on it, thus achieving satisfactory performance. However, our method achieves superior performance over supervised methods on the LSRW and MIT5K datasets, proving that our method is capable of generalizing to unseen scenes.

\subtitle{Backlit Image Enhancement.} For the backlit image enhancement task, we compare our method with the existing low-light image enhancement methods as well as two unsupervised backlit image enhancement methods CLIP-LIT~\cite{CLIPLIT} and RAVE~\cite{RAVE}. As reported in Table~\ref{tab: BIE_compare}, our method achieves the best performance in terms of both distortion and perceptual metrics compared with all competitors on the paired and unpaired benchmarks, demonstrating the superiority of our method in preserving visual fidelity.
\begin{table}[!t]
  \centering
  \caption{Quantitative comparisons on the BAID~\cite{BAID} and Backlit300~\cite{CLIPLIT} datasets for BIE. The best results are highlighted in \textbf{bold} and the second-best are in \underline{underlined}.}
  \large
    \resizebox{\linewidth}{!}{
    \begin{tabular}{l|ccc|cc}
    \toprule
    \multirow{2}[4]{*}{Methods} & \multicolumn{3}{c|}{BAID~\cite{BAID}} & \multicolumn{2}{c}{Backlit300~\cite{CLIPLIT}} \\
    \cmidrule{2-6} & PSNR $\uparrow$ & SSIM $\uparrow$ & LPIPS $\downarrow$ & NIQE $\downarrow$ & CLIPIQA $\uparrow$ \\
    \midrule
    RetinexNet~\cite{RetinexNet} & 15.851 & 0.781 & 0.229 & 4.036 & 0.441 \\
    KinD++~\cite{KinD++} & 18.849 & 0.824 & 0.194 & \underline{3.385} & 0.463 \\
    UHDFour~\cite{UHD_ICLR} & 18.537 & 0.673 & 0.348 & 4.839 & 0.366 \\
    PyDiff~\cite{PyDiff} & 19.881 & 0.836 & 0.224 & 3.744 & 0.473 \\
    AnlightenDiff~\cite{Anlightendiff} & 17.727 & 0.724 & 0.331 & 4.134 & 0.389 \\
    Reti-Diff~\cite{Retidiff} & 15.604 & 0.821 & 0.249 & 4.255 & 0.441 \\
    \midrule
    Zero-DCE~\cite{Zero-DCE} & 19.915 & 0.849 & 0.175 & 3.608 & 0.488 \\
    EnlightenGAN~\cite{EnlightenGAN} & 16.455 & 0.813 & 0.244 & 3.595 & 0.455 \\
    RUAS~\cite{RUAS}  & 9.972 & 0.612 & 0.535 & 6.225 & 0.349 \\
    SCI~\cite{SCI}   & 12.816 & 0.733 & 0.335 & 4.271 & 0.444 \\
    PairLIE~\cite{PairLIE} & 16.188 & 0.778 & 0.277 & 4.077 & 0.440 \\
    NeRCo~\cite{NeRCo} & 19.142 & 0.811 & 0.267 & 3.858 & 0.416 \\
    CLIP-LIT~\cite{CLIPLIT} & \underline{21.705} & \underline{0.862} & \underline{0.151} & 3.711 & 0.464 \\
    FourierDiff~\cite{FourierDiff} & 17.024 & 0.781 & 0.248 & 4.605 & 0.413 \\
    QuadPrior~\cite{QuadPrior} & 17.203 & 0.747 & 0.388 & 5.289 & 0.460 \\
    RAVE~\cite{RAVE}  & 20.086 & 0.836 & 0.181 & 3.625 & 0.491 \\
    LightenDiff~\cite{Lightendiffusion} & 20.985 & 0.838 & 0.215 & 3.559 & \underline{0.499} \\
    AGLLDiff~\cite{AGLLDiff} & 14.153 & 0.721 & 0.380 & 5.828 & 0.342 \\
    ZeroIDIR (Ours)  & \textbf{21.753} & \textbf{0.871} & \textbf{0.133} & \textbf{3.070} & \textbf{0.563} \\
    \bottomrule
    \end{tabular}}
  \label{tab: BIE_compare}%
\end{table}%
\begin{table*}[!t]
  \centering
  \caption{Quantitative comparisons on the MSEC~\cite{MSEC} and SICE~\cite{SICE} datasets for MEC. The best results are highlighted in \textbf{bold} and the second-best are in \underline{underlined}. `SL' and `UL' denote the methods as supervised and unsupervised, respectively.}
    \Large
  \resizebox{\linewidth}{!}{
    \begin{tabular}{c|l|c|ccc|ccc|ccc|ccc}
    \toprule
    \multirow{3}[6]{*}{Type} & \multirow{3}[6]{*}{Methods} & \multirow{3}[6]{*}{Reference} & \multicolumn{6}{c|}{MSEC~\cite{MSEC}} & \multicolumn{6}{c}{SICE~\cite{SICE}} \\
    \cmidrule{4-15} & & & \multicolumn{3}{c|}{Under} & \multicolumn{3}{c|}{Over} & \multicolumn{3}{c|}{Under} & \multicolumn{3}{c}{Over} \\
     & & & PSNR $\uparrow$ & SSIM $\uparrow$ & LPIPS $\downarrow$ & PSNR $\uparrow$ & SSIM $\uparrow$ & LPIPS $\downarrow$ & PSNR $\uparrow$ & SSIM $\uparrow$ & LPIPS $\downarrow$ & PSNR $\uparrow$ & SSIM $\uparrow$ & LPIPS $\downarrow$ \\
    \midrule
    \multirow{7}[2]{*}{SL} 
    & MSEC~\cite{MSEC} & CVPR' 21 & 20.582 & 0.818 & 0.188 & 20.016 & 0.827 & 0.177 & 17.473 & 0.565 & 0.310 & 15.618 & 0.565 & 0.377 \\
    & CMEC~\cite{CMEC}  & BMVC' 21 & 21.600 & 0.854 & 0.153 & 21.461 & 0.839 & 0.176 & 16.005 & 0.490 & 0.331 & 15.134 & 0.621 & 0.303 \\
    & FECNet~\cite{FECNet} & ECCV' 22 & 22.827 & 0.852 & 0.179 & 23.035 & 0.867 & 0.158 & 15.215 & 0.467 & 0.303 & 16.335 & 0.634 & \underline{0.272} \\
    & IAT~\cite{IAT}   & BMVC' 22 & 21.374 & 0.818 & 0.192 & 21.216 & 0.847 & 0.160 & 14.751 & 0.398 & 0.387 & 15.768 & 0.611 & 0.329 \\
    & MSLT~\cite{MSLT}  & WACV' 24 & 22.355 & 0.842 & 0.172 & 22.007 & 0.849 & 0.150 & 16.252 & 0.500 & 0.309 & 15.870 & 0.612 & 0.326 \\
    & CoTF~\cite{CoTF}  & CVPR' 24 & \textbf{23.407} & \textbf{0.863} & \underline{0.144} & \textbf{23.479} & \textbf{0.879} & \underline{0.124} & 16.524 & 0.516 & 0.337 & \underline{16.840} & \underline{0.638} & 0.288 \\
    & SLOT~\cite{Exposure-slot} & CVPR' 25 & \underline{23.127} & \underline{0.862} & \textbf{0.141} & \underline{23.213} & \underline{0.877} & \textbf{0.121} & 16.048 & 0.490 & 0.329 & 16.673 & 0.622 & 0.285 \\
    \midrule
    \multirow{7}[2]{*}{UL}
    & Exposure~\cite{Exposure} & ToG' 18 & 19.028 & 0.777 & 0.220 & 17.627 & 0.784 & 0.235 & 16.527 & 0.557 & 0.320 & 13.945 & 0.571 & 0.372 \\
    & DUAL~\cite{Dual}  & CFG' 19 & 16.879 & 0.727 & 0.291 & 15.838 & 0.701 & 0.277 & 14.655 & 0.476 & 0.417 & 11.831 & 0.472 & 0.385 \\
    & ExCNet~\cite{ExCNet} & MM' 19 & 13.470 & 0.699 & 0.274 & 12.800 & 0.683 & 0.295 & 15.957 & 0.574 & 0.322 & 11.881 & 0.499 & 0.421 \\
    & PEC~\cite{PEC}   & Arxiv' 22 & 16.371 & 0.758 & 0.226 & 15.117 & 0.729 & 0.254 & 18.111 & 0.611 & 0.305 & 14.837 & 0.618 & 0.293 \\
    & PSENet~\cite{Psenet} & WACV' 23 & 19.385 & 0.801 & 0.210 & 18.939 & 0.806 & 0.183 & \underline{18.469} & \underline{0.634} & \underline{0.239} & 13.366 & 0.571 & 0.291 \\
    & UEC~\cite{UEC}   & ECCV' 24 & 19.013 & 0.795 & 0.199 & 19.054 & 0.807 & 0.177 & 16.835 & 0.589 & 0.333 & 15.514 & 0.603 & 0.280 \\
    & ZeroIDIR (Ours)  & - & 20.573 & 0.805 & 0.193 & 21.673 & 0.839 & 0.145 & \textbf{18.573} & \textbf{0.659} & \textbf{0.215} & \textbf{16.975} & \textbf{0.661} & \textbf{0.259} \\
    \bottomrule
    \end{tabular}}
  \label{tab: MEC_compare}
\end{table*}
\begin{figure*}[!t]
    \centering
    \includegraphics[width=\linewidth]{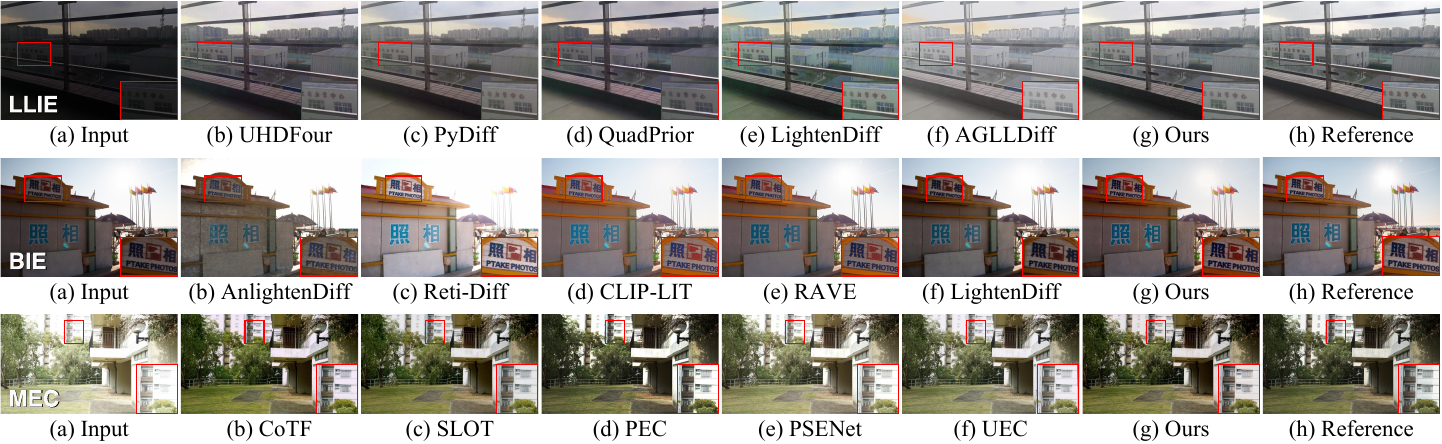}
    \caption{Qualitative comparison on the LSRW~\cite{R2RNet} (row 1), BAID~\cite{BAID} (row 2), and SICE~\cite{SICE} (row 3) test sets for low-light image enhancement (LLIE), backlit image enhancement (BIE), and multiple exposure correction (MEC), respectively.}
    \label{fig: visual_compare}
\end{figure*}

\subtitle{Multiple Exposure Correction.} For the multiple exposure correction task, we compare our method with existing state-of-the-art supervised methods, including MSEC~\cite{MSEC}, CMEC~\cite{CMEC}, FECNet~\cite{FECNet}, IAT~\cite{IAT}, MSLT~\cite{MSLT}, CoTF~\cite{CoTF}, and SLOT~\cite{Exposure-slot}, and unsupervised methods, including Exposure~\cite{Exposure}, DUAL~\cite{Dual}, ExCNet~\cite{ExCNet}, PEC~\cite{PEC}, PSENet~\cite{Psenet}, and UEC~\cite{UEC}. For fair comparisons, we adopt the released weights pre-trained on the MSEC dataset of supervised methods for evaluation. As reported in Table~\ref{tab: MEC_compare}, our method performs competitively on the over-exposed correction part of the MSEC dataset and even outperforms several supervised methods trained on it, such as IAT and CMEC. Supervised methods achieve better performance on the MSEC dataset, while presenting limited generalization capability to the SICE dataset. In contrast, unsupervised approaches demonstrate better generalization across datasets, where our method achieves the best performance on the SICE dataset in terms of both distortion and perceptual metrics, highlighting our robustness and adaptability to diverse illumination conditions.

\subsection{Qualitative Comparison}\label{subsec: Qualitative_Comparison}
We present visual comparisons of our method and competitive methods on the LSRW~\cite{R2RNet}, BAID~\cite{BAID}, and SICE~\cite{SICE}test sets for low-light image enhancement, backlit image enhancement, and multiple exposure correction, respectively. As shown in Fig.~\ref{fig: visual_compare}, previous methods yield results with under-/over-exposed, color distortion, texture loss, or noise amplification. In contrast, our method properly corrects global and local illumination, reconstructs finer details, presents vivid color, and suppresses noise, resulting in visually pleasing results.

\subsection{Ablation Study}
We conduct a series of ablation studies to validate the impact of different component choices. We use the implementation described in Sec.~\ref{subsec:experimental_settings} for training, and quantitative results on the LOL~\cite{RetinexNet} and the over-exposed correction part of the SICE~\cite{SICE} datasets are reported in Table~\ref{tab: ablation_1}-\ref{tab: ablation_2}.
\begin{table}[!t]
  \centering
  \caption{Quantitative results of ablation studies about the proposed AGCM. `w/o' denotes without. Please refer to the text for details.}
  \Large
  \resizebox{\linewidth}{!}{
    \begin{tabular}{l|ccc|ccc}
    \toprule
    \multirow{2}[4]{*}{Method} & \multicolumn{3}{c|}{LOL~\cite{RetinexNet}} & \multicolumn{3}{c}{SICE-over~\cite{SICE}} \\
    \cmidrule{2-7} & PSNR $\uparrow$ & SSIM $\uparrow$ & LPIPS $\downarrow$ & PSNR $\uparrow$ & SSIM $\uparrow$ & LPIPS $\downarrow$ \\
    \midrule
    GC$_{\gamma=\{0.1,5.0\}}$ & 13.404 & 0.339 & 0.625 & 13.923 & 0.527 & 0.316 \\
    GC$_{\gamma=\{0.2,5.5\}}$ & 16.805 & 0.441 & 0.501 & 13.937 & 0.509 & 0.328 \\
    GC$_{\gamma=\{0.3,6.0\}}$ & 16.332 & 0.508 & 0.416 & 13.890 & 0.465 & 0.360 \\
    \midrule
    `w/o' Retinex & 18.378 & 0.631 & 0.397 & 14.434 & 0.595 & 0.344 \\
    `w/o' $\mathcal{L}_{hic}$ & 18.406 & 0.528 & 0.376 & 14.032 & 0.572 & 0.305 \\
    \midrule
    Default & 19.599 & 0.535 & 0.351 & 14.506 & 0.574 & 0.293 \\
    \bottomrule
    \end{tabular}}
  \label{tab: ablation_1}%
\end{table}%
\begin{figure}[!t]
    \centering
    \includegraphics[width=\linewidth]{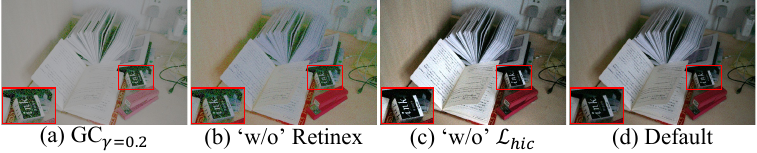}
    \caption{Visual results of the ablation study about our AGCM.}
    \label{fig: visual_ablation_1}
\end{figure}

\subtitle{Effectiveness of AGCM.} To evaluate the effectiveness of the AGCM, we selectively modify or remove its core components for evaluation. First, we compare AGCM with the baseline approach that applies traditional Gamma correction (GC) directly to the illumination map to generate the illumination-corrected results as $I^{'}_{d} = {\mathbf{L}_{d}}^{\gamma} \odot \mathbf{R}_{d}$, we set $\gamma=\{0.1, 0.2, 0.3\}$ for the LOL dataset and $\gamma=\{5.0, 5.5, 6.0\}$ for the SICE dataset. As reported in rows 1-3 of Table~\ref{tab: ablation_1}, it is difficult to select a universal gamma value that works across diverse conditions, and the global nature of GC often causes inflexible exposure correction, as shown in Fig.~\ref{fig: visual_ablation_1}(a). Then, we remove the Retinex decomposition and apply AGCM directly to the entire image. Without isolating illumination from reflectance, it would amplify the residual noise and cause color distortions, as shown in Fig.~\ref{fig: visual_ablation_1}(b). Moreover, we ablate the proposed $\mathcal{L}_{hic}$, which is designed to constrain the corrected illumination distribution to align with well-illuminated statistics. As reported in row 5 of Table~\ref{tab: ablation_1}, without this regularization leads to overall performance degradation and causes unstable brightness correction as shown in Fig.~\ref{fig: visual_ablation_1}(c). These results collectively demonstrate that each component of AGCM plays a critical role in achieving adaptive illumination correction while preserving content information and avoiding noise amplification, providing reliable input for subsequent reconstruction.

\begin{figure}[!t]
    \centering
    \includegraphics[width=\linewidth]{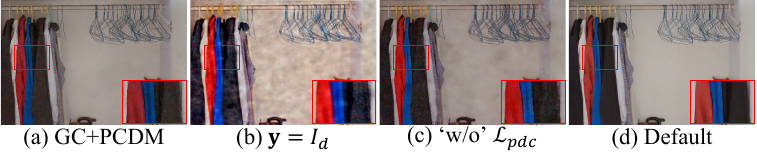}
    \caption{Visual results of the ablation study about our PCDM.}
    \label{fig: visual_ablation_2}
\end{figure}

\subtitle{Effectiveness of PCDM.} We further conduct experiments to validate the effectiveness of the proposed PCDM. We first replace the illumination-corrected result obtained from our AGCM with the traditional gamma-corrected image, i.e., row 2 of Table~\ref{tab: ablation_1}, as the intermediate perturbed noisy state for PCDM. As reported in row 1 of Table~\ref{tab: ablation_2}, our PCDM still yields noticeable improvements over the baseline approach, demonstrating its strong generative and denoising capabilities in restoring fine-grained details and suppressing noise even when starting from the less reliable intermediate state. Moreover, instead of using the illumination-corrected image as the condition for the diffusion model, we revert to the original low-quality degraded image like most previous methods~\cite{Retidiff, Anlightendiff, Lightendiffusion}. This modification introduces exposure bias into the learning process, as the model is forced to jointly address both illumination inconsistency and detail degradation, resulting in the restored result suffering from insufficient visual fidelity and unstable exposure correction, as shown in Fig.~\ref{fig: visual_ablation_2}(b). In contrast, using the illumination-corrected image as the condition encourages the model to concentrate on detail reconstruction and denoising, thus generating better results as shown in Fig.~\ref{fig: visual_ablation_2}(d). Lastly, we ablate the perturbed diffusion consistency loss $\mathcal{L}_{pdc}$, which is designed to enforce trajectory consistency between the restored image and the illumination-corrected intermediate state. As reported in row 3 of Table~\ref{tab: ablation_2}, without $\mathcal{L}_{pdc}$ causes the diffusion trajectory drifts to degrade overall performance, resulting in unexpected artifacts and unsatisfactory visual quality as shown in Fig.~\ref{fig: visual_ablation_2}(c).
\begin{table}[!t]
  \centering
  \caption{Quantitative results of ablation studies about the proposed PCDM. `w/o' denotes without. Please refer to the text for details.}
  \Large
  \resizebox{\linewidth}{!}{
    \begin{tabular}{l|ccc|ccc}
    \toprule
    \multirow{2}[4]{*}{Method} & \multicolumn{3}{c|}{LOL~\cite{RetinexNet}} & \multicolumn{3}{c}{SICE-over~\cite{SICE}} \\
    \cmidrule{2-7} & PSNR $\uparrow$ & SSIM $\uparrow$ & LPIPS $\downarrow$ & PSNR $\uparrow$ & SSIM $\uparrow$ & LPIPS $\downarrow$ \\
    \midrule
    GC + PCDM & 18.369 & 0.716 & 0.386 & 15.562 & 0.569 & 0.296 \\
    \midrule
    $\mathbf{y} = I_{d}$ & 17.571 & 0.619 & 0.432 & 15.520 & 0.598 & 0.304 \\
    `w/o' $\mathcal{L}_{pdc}$ & 19.780 & 0.789 & 0.197 & 16.012 & 0.601 & 0.291 \\
    \midrule
    Default & 20.874 & 0.811 & 0.167 & 16.975 & 0.661 & 0.259 \\
    \bottomrule
    \end{tabular}}
  \label{tab: ablation_2}%
\end{table}%

\section{Conclusion}
We have presented ZeroIDIR, a zero-reference diffusion-based framework for illumination degradation image restoration trained solely on low-quality degraded images, which decouples the restoration process into illumination correction and diffusion-based reconstruction. Technically, we design an adaptive gamma correction module to estimate spatially varying gamma maps for exposure correction to generate illumination-corrected results, where a histogram-guided illumination correction loss is proposed to constrain the corrected illumination distribution to align with that observed in natural scenes. Subsequently, we propose a perturbed consistency diffusion model to treat the illumination-corrected image as the intermediate noisy state for diffusion processes, where a perturbed diffusion consistency loss is further proposed to facilitate detail reconstruction and noise suppression. Extensive experiments demonstrate that our method outperforms existing state-of-the-art competitors both quantitatively and qualitatively.

\newpage
\noindent\textbf{Acknowledgments.} This work was supported by the National Natural Science Foundation of China (NSFC) under grants No.625B2123, No.62372091, No.U23B2013, and No.62276176, and the Hainan Province Science and Technology Plan Project under grant ZDYF2024(LALH)001.
{
    \small
    \bibliographystyle{ieeenat_fullname}
    \bibliography{main}
}

\end{document}